\documentclass{article}

\usepackage{PRIMEarxiv}

\usepackage[utf8]{inputenc} % allow utf-8 input
\usepackage[T1]{fontenc}    % use 8-bit T1 fonts
\usepackage{hyperref}       % hyperlinks
\usepackage{url}            % simple URL typesetting
\usepackage{booktabs}       % professional-quality tables
\usepackage{amsfonts}       % blackboard math symbols
\usepackage{nicefrac}       % compact symbols for 1/2, etc.
\usepackage{microtype}      % microtypography
\usepackage{lipsum}
\usepackage{fancyhdr}       % header
\usepackage{graphicx}       % graphics
\usepackage{hyphenat}
\graphicspath{{media/}}     % organize your images and other figures under media/ folder

\hypersetup{hidelinks}

\usepackage{pifont}
\newcommand{\cmark}{\ding{51}}
\newcommand{\xmark}{\ding{55}}

\usepackage{amsmath,amssymb}
\usepackage{chngcntr}
\usepackage{algorithmic}
\usepackage{algorithm}
\usepackage{array}
\usepackage[caption=false,font=normalsize,labelfont=sf,textfont=sf]{subfig}
\usepackage{textcomp}
\usepackage{verbatim}
\usepackage{tabularx}
\usepackage{natbib}
\usepackage{bm}
\usepackage{wrapfig}
\usepackage{caption}
\hypersetup{
    colorlinks=true,
    citecolor=blue,
    linkcolor=blue,
    urlcolor=blue
}
\title{\bf
Arrive and Survive: Scaling Safe Goal-Conditioned Policy Learning from One-Bit Failure Signals
}

\author{
  Guopeng Li,\ Yiyang Duan,\ Chengcheng Xu\\
  Southeast University \\
  Nanjing, China\\
  \texttt{\{guopengli5, duanyiyang, xuchengcheng\}@seu.edu.cn} \\
   \And
  Yiru Jiao\\
  2030 Lab, Yinwang Intel. Tech. Co. Ltd. \\
  Shanghai, China\\
  \texttt{yiru.jiao@studyinger.space} \\
}

\begin{document}
\maketitle

\begin{abstract}
Contrastive reinforcement learning (CRL) scales effectively in goal-conditioned tasks by casting policy learning into a self-supervised contrastive objective. However, in a failure-terminated Markov decision process, established CRL considers pre-failure future goals only when constructing positive samples, without accounting for the probability mass removed by failure termination. Our theoretical analysis shows that this omission induces a systematic overestimation bias in goal-reaching values. Consequently, near-failure trajectories provide disproportionately strong supervision of success despite retaining little future occupancy. Unsafe actions can thereby be reinforced through catastrophic failure bootstrapping, leading to failed policy learning and unsustainable goal-reaching behaviours. To address this problem, we introduce two minimal yet strong corrections: mass-weighted InfoNCE corrects the overweighting of short surviving futures in critic learning, and a log-survival-mass score restores the missing survival mass in policy optimization. The resulting method, Safe Contrastive Reinforcement Learning (Safe-CRL), requires only the one-bit signal provided by failure termination to scale safe goal-conditioned policy learning. Across twelve failure-prone robot navigation and locomotion tasks, Safe-CRL consistently improves survival and substantially outperforms the Scaling-CRL baseline in goal-reaching performance. Additionally, deep Safe-CRL policies exhibit complex failure-avoidance behaviours. This study completes the CRL theory under failure termination and provides a scalable safe RL framework. The code is available via \hyperlink{https://github.com/RomainLITUD/safe-crl}{https://github.com/RomainLITUD/safe-crl}.
\end{abstract}
% keywords can be removed

\begin{figure}[h!]
\begin{center}
%\framebox[4.0in]{$\;$}
\includegraphics[width=0.81\columnwidth]{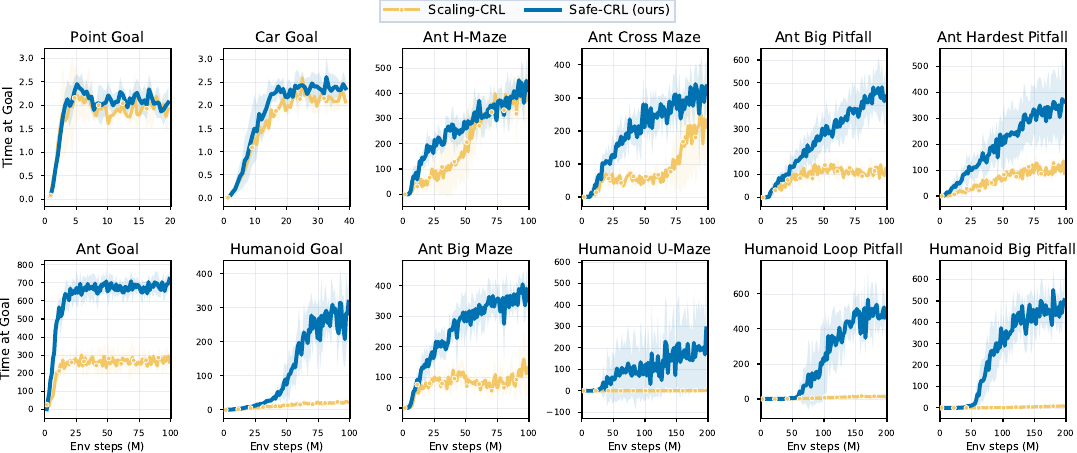}
\end{center}
\caption{Comparison of time at goal between Scaling-CRL and the proposed Safe-CRL method in failure-prone navigation and locomotion tasks (mean $\pm$ standard deviation over five random seeds). Both methods use 64-layer actor and critic networks (8-layer for Point Goal and Car Goal). Safe-CRL achieves higher or on-par mean time at goal across all tasks, with pronounced gains in nine.}
\label{fig: open-figure}
\end{figure}

\section{Introduction}
\label{sec: intro}

Scaling model capacity has become an important route toward more capable reinforcement learning (RL) policies~\cite{schwarzer2023bigger,lee2025simba}. In goal-conditioned RL, contrastive reinforcement learning (CRL) is particularly promising. It turns sparse goal-reaching reward into dense self-supervision through goal relabelling~\cite{eysenbach2022contrastiverl,andrychowicz2017her}, and learns to reach commanded goals via a contrastive objective. Recent work on Scaling-CRL has shown that this approach can effectively exploit deeper actor and critic networks to improve performance, establishing CRL as a scalable policy learning framework~\cite{wang2025scalingcrl}.

In practice, many goal-conditioned tasks involve \textit{failure termination} -- unsafe events like falls or collisions immediately end the episode. For example, robot locomotion requires reaching the goal position without falls~\cite{berseth2021smirl, jain2021pixels}; navigation and autonomous driving require chasing moving goals while avoiding collisions~\cite{belkhouche2006line, bansal2019chauffeurnet}. In these safety-critical settings, failure prevents the goal from being reached or terminates the episode shortly after the agent reaches the goal. Therefore, a safe goal-conditioned policy needs to learn how to arrive and to survive in a scalable manner. 
A desirable solution should preserve the self-supervised scalability of CRL without introducing additional safety annotations or task-specific cost signals. Ideally, learning should rely only on the one-bit failure signal already provided by termination.

However, failure termination interacts with the established CRL method subtly: failure truncates the future states available for relabelling, but established CRL renormalizes the remaining pre-failure future goals back to unit mass~\cite{eysenbach2022contrastiverl, wang2025scalingcrl}. For example, from the same state, consider two actions that lead to similar observed pre-failure future goals: one is followed by a long stable trajectory, whereas the other leads to failure quickly after reaching the goal. Relabelling the valid futures of each trajectory without accounting for their different trajectory lengths gives the two actions comparable positive supervision despite different safety performance. In other words, CRL preserves information about \textit{which} pre-failure future goals are reachable, but loses information about \textit{how much} discounted future-goal occupancy survives, which we term the \textit{survival mass}.

\begin{figure}[h!]
\centering
    \includegraphics[width=\linewidth]{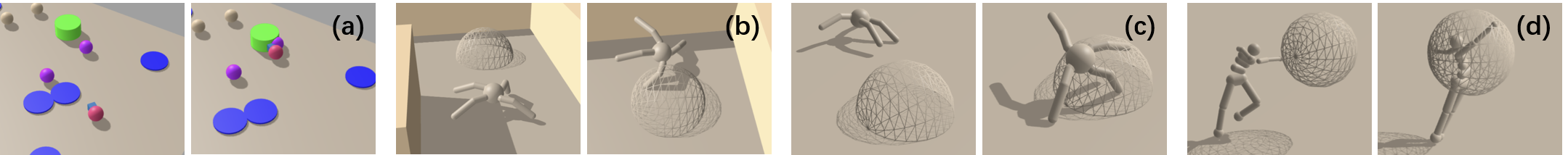}
    \caption{Examples of unsustainable goal reaching under Scaling-CRL. (a) The Point agent reaches the goal region but subsequently collides with a moving obstacle; (b) the Ant reaches the goal at high speed and overshoots into the wall; (c–d) locomotion agents reach the goal but fall immediately afterwards. These examples illustrate that successful short-term goal reaching does not necessarily lead to a stable trajectory under failure termination.}
    \label{fig: unsafe-reach}
\end{figure}

The missing survival mass induces a systematic bias in goal-reaching value and can have severe consequences. The first one is what we refer to as \textit{catastrophic failure bootstrapping}. In failure-prone tasks, the replay buffer is filled with short near-failure trajectories during the early stages of training. Repeatedly reinforcing such trajectories through self-supervision can trap learning in failure-prone behaviour. Another consequence is \textit{unsustainable goal reaching}: the agent aggressively reaches a commanded goal but collides, overshoots, or falls shortly afterwards, as illustrated in Figure~\ref{fig: unsafe-reach}. These two failure modes can substantially degrade CRL performance in safety-critical tasks.

The discussion above raises a critical research question: \textit{How should CRL account for failure termination to realize scalable and safe goal-conditioned policy learning, when only a one-bit failure signal is available?}

% The difficulty of building scalable safe CRL is subtle because failure termination changes with not only which future states are observed, but also how much future occupancy remains valid after an action. 

In this paper, we first identify and formalize the systematic overestimation bias in established CRL under failure termination. This bias can trap policy learning in unsafe, near-failure regions, or cause unsustainable goal reaching. To address this problem, we introduce \textit{mass-weighted InfoNCE} and a \textit{log-survival-mass correction}, yielding \textbf{Safe-CRL} as a minimal yet strong extension of Scaling-CRL~\cite{wang2025scalingcrl}. Safe-CRL substantially improves survival and goal-reaching performance in failure-prone environments while preserving Scaling-CRL's depth scalability.

\section{The missing survival mass in CRL}
\label{sec: gap}

We now formalize how failure termination absorbs discounted future-goal occupancy and explain why established CRL has an overestimation bias caused by discarding this information. We then show how the bias affects critic learning and actor optimization.

\subsection{Problem formulation and CRL preliminaries}
\label{sec: preliminaries}

We consider a goal-conditioned MDP with failure termination,
\begin{equation}
\mathcal{M} = \left( \mathcal{S}, \mathcal{A}, P, \rho_0, \gamma, \mathcal{G}, \phi, F \right),
\end{equation}
where $\mathcal{S}$ and $\mathcal{A}$ denote the state and action spaces, $P(s'\mid s,a)$ is the state transition kernel, $\rho_0$ is the initial-state distribution, and $\gamma\in(0,1)$ is the discount factor. The goal space is $\mathcal{G}$, and $\phi:\mathcal{S}\rightarrow\mathcal{G}$ maps a state to the achieved goal. A goal-conditioned policy $\pi(a\mid s,g)$ acts toward a commanded goal $g\in\mathcal{G}$. The binary function $F:\mathcal{S}\times\mathcal{A}\times\mathcal{S}\rightarrow\{0,1\}$ indicates whether a transition leads to a failure that immediately terminates the episode. We denote $T_f$ as the random variable representing the failure horizon measured from an anchor $(s_t, a_t)$.

CRL uses a contrastive critic together with a goal-conditioned actor and operates on \textit{anchor-goal pairs}. The critic learns a contrastive score $f_\theta(x,g)$ between a state-action anchor $x=(s,a)$ and a goal $g$. Positive goals are obtained by hindsight future-goal relabelling~\cite{eysenbach2022contrastiverl,andrychowicz2017her}: a horizon $H$ is sampled from a discounted geometric distribution $q_\gamma(h)=(1-\gamma)\gamma^{h-1}$ for $h\geq 1$, and the observed achieved goal $g_{t+H} = \phi(S_{t+H})$ is the positive goal for this anchor. Goals associated with other anchors in a mini-batch are considered negatives. Thus, the InfoNCE loss is used to train the critic to distinguish the positive goals from negative goals by learning higher scores. As the positive goal is actually reached from the anchor, this learning process is interpreted as maximizing the likelihood of reachability in CRL~\cite{wang2025scalingcrl}.

At the population optimum, the contrastive score represents a positive-to-negative density ratio. Let $p^+(g\mid x)$ denote the positive goal distribution from anchor $x$ and consider an anchor-independent negative proposal $p^-(g)$. The contrastive score estimates~\cite{eysenbach2022contrastiverl}
\begin{equation}
f_\theta(x,g) \rightarrow \log \frac{p^{+}(g\mid x)}{p^{-}(g)} + c(x),
\end{equation}
where $c(x)$ is an anchor-dependent offset. Because $x=(s,a)$, row-wise InfoNCE alone does not make scores directly comparable across candidate actions. CRL therefore uses score calibration to constrain this offset~\cite{eysenbach2022contrastiverl}. Under the calibrated score convention, for a fixed commanded goal $g$, the negative proposal is action-independent, and actor optimization reduces to maximizing the conditional positive-goal likelihood, where $\bar f_\theta$ denotes the calibrated contrastive score:
\begin{equation}
    \arg\max_a \bar f_\theta(x,g) = \arg\max_a \log p^{+}(g\mid x).
    \label{eq: actor objective}
\end{equation}

\subsection{Overestimation bias in established CRL}

Next, we show how failure termination and pre-failure-only relabelling in CRL create a systematic overestimation bias. Following CRL~\cite{eysenbach2022contrastiverl}, throughout the occupancy analysis below, $\pi$ denotes the replay-induced trajectory distribution underlying contrastive future-goal sampling; when replay contains trajectories collected under different commanded goals, it is understood as the corresponding goal-marginalized replay-induced trajectory distribution. For an MDP with infinite horizon (no failure termination), Eysenbach et al.~\cite{eysenbach2022contrastiverl} established the equivalence between the goal-reaching action-value function (Q-function) and the so-called \textit{discounted state occupancy}, defined by
\begin{equation}
\mu_{\infty}^{\pi}(g\mid x) = \sum_{h=1}^{\infty} q_\gamma(h) p^{\pi} \left( \phi(s_{t+h})=g \mid x_t=x \right), 
\label{eq: occupancy}
\end{equation}
Proposition 1 of Eysenbach et al.~\cite{eysenbach2022contrastiverl} shows
\begin{equation}
    Q_g^\pi(s,a) = \mu_{\infty}^{\pi}(g\mid x).
\end{equation} 
The occupancy measure of Eq.~\eqref{eq: occupancy} forms a valid positive goal distribution with unit probability mass for Eq.~\eqref{eq: actor objective}, and maximizing the contrastive score is equivalent to maximizing the goal-reaching Q-function. 

Under failure termination, however, post-failure states are no longer valid future goals. In such settings, we propose:
\paragraph{Proposition 1 (Q-function is equivalent to failure-aware discounted future-goal occupancy)}
Let $T_f$ denote the failure horizon. We define the failure-aware discounted future-goal occupancy as
\begin{equation}
\mu_f^\pi(g\mid x) = \sum_{h=1}^{\infty} q_\gamma(h) p^\pi \left( \phi(s_{t+h})=g,\ T_f > h \mid x_t=x \right), 
\label{eq: failure-aware occupancy}
\end{equation}
Then we have the following equivalence:
\begin{equation}
Q_g^\pi(s,a)=\mu_f^\pi(g\mid x).
\end{equation}

The proof is provided in Appendix~\ref{app:proof_prop1}. Unlike the infinite-horizon occupancy in Eq.~\eqref{eq: occupancy}, $\mu_f^\pi$ is \textit{NOT} a valid goal distribution. It is generally a sub-probability measure because failure removes part of the discounted future occupancy. Its total mass is
\begin{equation}
Z^\pi(x)
\triangleq \int_{\mathcal G} \mu_f^\pi(g\mid x) \,dg 
=\sum_{h=1}^{\infty} q_\gamma(h) \Pr^\pi \left( T_f>h \mid x\right)
=\mathbb{E}_{H\sim q_\gamma}\left[ \Pr^\pi \left( T_f>H \mid x\right)\right] \leq 1.
\label{eq: survival mass}
\end{equation}
We call $Z^\pi(x)$ \textbf{survival mass}. Intuitively, $Z^\pi(x)$ is the probability that a future horizon sampled from $q_\gamma$ still lies before failure.

However, established CRL uses only the valid pre-failure futures as positives and renormalizes their remaining occupancy to unit mass before entering the contrastive objective. Failure therefore changes which futures remain, but the amount of discounted occupancy removed by failure is lost after normalization. Let $\tau$ denote the observed future trajectory from anchor $x_t=x$, and let $L_\tau$ be the number of valid future states before failure. The realized failure-aware occupancy and its realized survival mass are
\begin{equation}
    \hat{\mu}_\tau(g \mid x) = \sum_{h=1}^{L_{\tau}} q_\gamma(h) \textbf{1}\{\phi(s_{t+h}) = g\}, \quad \hat Z_{\tau} = \sum_{h=1}^{L_{\tau}} q_\gamma(h) = 1-\gamma^{L_{\tau}}.
\end{equation}
CRL samples positive goals from this realized trajectory after normalizing its valid future occupancy, so the trajectory-level positive distribution is $\hat\mu_\tau(g\mid x)/\hat Z_\tau$.
By construction, on average,
\begin{equation}
    \mathbb{E}_{\tau\mid x}[\hat{\mu}_\tau(g\mid x)] = \mu_f^\pi(g\mid x), \quad \mathbb{E}_{\tau\mid x}[\hat Z_\tau] = Z^\pi(x).
\end{equation}
The population positive-goal distribution is
\begin{equation}
    p^+_{\text{CRL}}(g\mid x) = \mathbb{E}_{\tau\mid x} \left[ \dfrac{\hat{\mu}_\tau(g\mid x)}{\hat Z_\tau}\right].
    \label{eq: traj-wise norm}
\end{equation}

Because $Z_\tau \in (0, 1]$, if we denote by $\hat{Q}^\pi_{\text{CRL},g}(s, a)$ the goal-reaching value implicitly represented by CRL, we have:

\begin{equation}
\boxed{
   \hat Q^\pi_{\text{CRL}, g}(s, a) = p^+_{\text{CRL}}(g\mid x) = \mathbb{E}_{\tau\mid x} \left[ \dfrac{\hat{\mu}_\tau(g\mid x)}{\hat Z_\tau}\right] \geq \mathbb{E}_{\tau\mid x}[\hat{\mu}_\tau(g\mid x)] = \mu_f^\pi(g\mid x) = Q^\pi_g(s, a).
}
   \label{eq: bias}
\end{equation}

Eq.~\eqref{eq: bias} establishes a \textit{non-negative overestimation bias}, while the trajectory-level amplification factor $1/\hat Z_\tau$ increases as $L_\tau$ decreases. This theoretical analysis provides a mechanism that leads to catastrophic failure bootstrapping and unsustainable goal reaching.

\subsection{Biased critic and actor learning in CRL}
\label{sec: bias role}

\begin{figure}[h!]
\centering
\includegraphics[width=0.9\columnwidth]{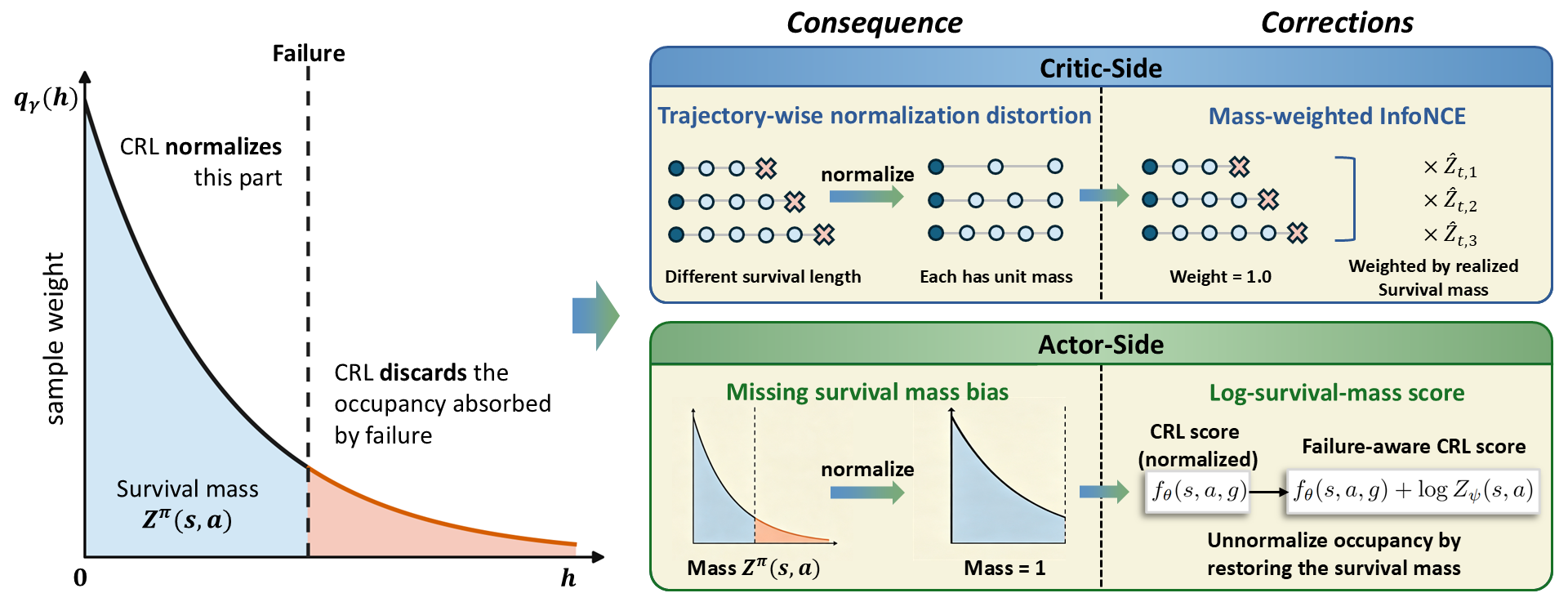}
    \caption{Failure termination removes part of the discounted future-goal occupancy. Established CRL normalizes the remaining pre-failure futures, whereas Safe-CRL corrects the trajectory-wise normalization distortion in critic learning and restores the missing survival mass in actor updates.}
    \label{fig: concept}
\end{figure}

Figure~\ref{fig: concept} conceptualizes two distinct consequences of the same normalization operation. First, normalizing each realized trajectory before averaging \textit{distorts} the conditional goal distribution learned by the critic. Second, even after this distortion is corrected, the contrastive objective in Eq.~\eqref{eq: actor objective} still represents a normalized distribution and therefore omits its action-dependent total survival mass.

\begin{itemize}
    \item \textbf{On critic learning:} Normalizing each trajectory to unit mass before averaging gives short trajectories too much relative weight compared with long trajectories. Therefore, the positive distribution learned by established CRL generally differs from the normalized population failure-aware occupancy,
    \begin{equation}
        p^+_{\text{CRL}}(g\mid x) \neq \bar{\mu}_f^\pi(g\mid x) \triangleq \dfrac{\mu_f^\pi(g\mid x)}{Z^\pi(x)}.
    \end{equation}
    Short near-failure trajectories are over-weighted relative to long safe trajectories, distorting the conditional goal distribution learned by the critic.

    \item \textbf{On actor optimization:} Correcting the critic-side distortion is not sufficient. Contrastive learning can recover only the normalized average conditional occupancy $\bar\mu_f^\pi$, whereas the true failure-aware goal-conditioned value factorizes as
    \begin{equation}
        Q^\pi_g(s, a)\triangleq \mu_f^\pi(g\mid x) = Z^\pi(x) \bar{\mu}_f^\pi(g\mid x).
        \label{eq: multiplier}
    \end{equation}
    The missing factor is exactly the action-dependent survival mass $Z^\pi(x)$. Actions with similar conditional goal distributions but very different survival masses can receive similar contrastive scores.
\end{itemize}

In summary, trajectory-wise normalization causes the critic-side trajectory-wise normalization distortion, while contrastive normalization removes the actor-side survival mass required for correct goal-conditioned action evaluation. We next rigorously correct the overestimation bias in CRL by restoring $Z^\pi(x)$.

\section{Safe-CRL by restoring the survival mass}
\label{sec: methodology}
The two effects identified above suggest two corresponding corrections, as illustrated in the right part of Figure~\ref{fig: concept}.

\subsection{Critic-side correction: mass-weighted InfoNCE}
Since each realized trajectory is normalized by its realized survival mass $\hat Z_i$ when constructing positive goals, we restore its relative contribution by weighting the complete InfoNCE row by the same $\hat Z_i$. Let $\ell_i^{\mathrm{NCE}}$ denote the complete row-wise InfoNCE loss~\cite{oord2018cpc,bortkiewicz2025jaxgcrl} for anchor $i$. This gives the \textit{mass-weighted InfoNCE} (MW-InfoNCE):
\begin{equation}
\mathcal L_{\mathrm{MW\text{-}NCE}}
=\frac{1}{B}\sum_{i=1}^{B}\hat Z_i\ell_i^{\mathrm{NCE}},
\label{eq: mw-infonce}
\end{equation}
where $B$ is the batch size. This gives the following proposition:

\paragraph{Proposition 2 (Mass-weighted InfoNCE corrects trajectory-wise normalization distortion)}
Eq.~\eqref{eq: mw-infonce} induces, at the population level, the normalized conditional distribution associated with the failure-aware occupancy,
\begin{equation}
p^+_{\text{W}}(g\mid x) = \dfrac{\mu_f^\pi(g\mid x)}{Z^\pi(x)} = \bar{\mu}_f^\pi(g\mid x).
\end{equation}

The proof is provided in Appendix~\ref{app:proof_prop2}. Intuitively, multiplying each row by $\hat Z_i$ cancels the trajectory-wise normalization in expectation.

\subsection{Actor-side correction: log-survival-mass correction}
Using the factorization in Eq.~\eqref{eq: multiplier}, once the critic recovers $\bar\mu_f^\pi$, the normalized contrastive score differs from the desired failure-aware occupancy score by exactly $-\log Z^\pi(x)$.

\paragraph{Corollary 1 (Restoring survival mass in actor scoring)}
Under the calibrated CRL score convention in Eq.~\eqref{eq: actor objective}, the optimal mass-weighted contrastive score satisfies
\begin{equation}
\bar f_W^*(x,g)= \log\dfrac{\bar{\mu}_f^\pi(g\mid x)}{p^-(g)} = \log \mu_f^\pi(g\mid x)- \log Z^\pi(x) - \log{p^-(g)}
\end{equation}
Because $\log{p^-(g)}$ is action-independent,
\begin{equation}
\arg\max_a\left[\bar f_W^*(x,g)+\log Z^\pi(x)\right]
=\arg\max_a\mu_f^\pi(g\mid x) = \arg\max_a Q^\pi_g(s,a).
\label{eq: our obj}
\end{equation}
The derivation is provided in Appendix~\ref{app:proof_corollary}. This corollary says that adding the $\log Z$ term restores the missing survival mass in the actor score and recovers the same action ranking as the failure-aware goal-conditioned Q-value.

\subsection{Implementation of corrections}
Safe-CRL implements these two corrections with a lightweight Z-encoder $Z_\psi(x)\in(0,1)$. For an anchor $x_t$, its realized survival mass $\hat Z_t$ weights the InfoNCE row as in Eq.~\eqref{eq: mw-infonce}. To train the Z-encoder, we independently sample $H_t\sim q_\gamma$ and use the binary survival label $Y_t=\mathbf{1}\{T_f>H_t\}$. If $H_t$ extends beyond the available trajectory without an observed failure, including time-limit truncation, we treat it as a survived sample and set $Y_t=1$; if failure is observed at or before $H_t$, $Y_t=0$. The Z-encoder is trained by
\begin{equation}
\mathcal{L}_{Z}(\psi)
=
-\mathbb{E}
\left[
Y_t \log Z_\psi(x_t)
+
(1-Y_t)\log\!\left(1-Z_\psi(x_t)\right)
\right].
\label{eq:z-encoder-loss}
\end{equation}
For the uncensored process, $\mathbb{E}[Y_t\mid\tau_t]=\hat Z_t$ and $\mathbb{E}[Y_t\mid x_t]=Z^\pi(x_t)$, so Eq.~\eqref{eq:z-encoder-loss} is a one-sample Monte Carlo estimator of the corresponding survival-mass BCE; see Appendix~\ref{app:z_population}.

For actor updates, we add the $\log Z$ term predicted by the Z-encoder to the actor objective:
\begin{equation}
f_{\mathrm{corrected}}(s,a,g)=f_\theta(s,a,g)+\log Z_\psi(s,a).
\end{equation}
In the implementation, we retain the log-sum-exp score regularization of Scaling-CRL to constrain anchor-dependent score offsets, and apply the log-$Z$ correction under the same score convention.

In practice, $\log Z_\psi$ is evaluated from the Z-encoder logit output using $\log\sigma(z_\psi)=-\operatorname{softplus}(-z_\psi)$ for numerical stability. During the actor update, the Z-encoder parameters are fixed while gradients through its action input are retained. All remaining Scaling-CRL components, including the ResNet-like network architecture, stay unchanged~\cite{wang2025scalingcrl}. We call the resulting method \textbf{Safe Contrastive Reinforcement Learning (Safe-CRL)}.

\paragraph{Why not model failure as an explicit future outcome?}
In principle, formulating failure as an absorbing outcome would preserve the missing mass. However, this changes the established CRL outcome space: failure is shared across many anchors and therefore requires special treatment in contrastive negative sampling, while it is not a valid commanded goal for actor learning. Safe-CRL instead retains the original achieved-goal space and restores the missing survival mass directly.

\section{Experiments}
\label{sec: exp}

\begin{figure}[h!]
\begin{center}
\includegraphics[width=0.75\columnwidth]{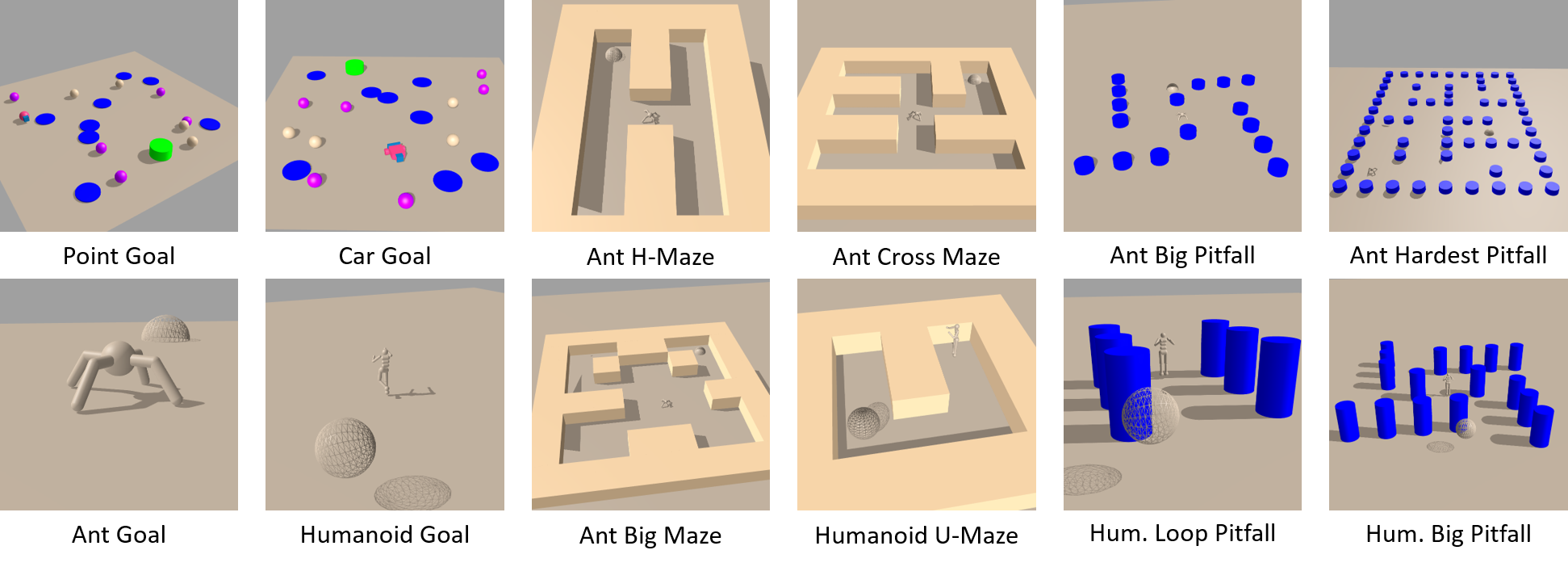}
\end{center}
\caption{Illustrations of the experimental environments. The goal is represented by a bright green cylinder (for Point and Car Goal) or a meshed sphere (for Ant and Humanoid locomotion tasks). The walls in Maze environments are concrete, while other obstacles, hazards (pitfalls), and moving gremlins (purple spheres in Point Goal and Car Goal) are non-colliding (ghost objects). Contacting maze walls is not a failure; entering a designated task-specific failure region triggers failure termination. The goal is randomly respawned only for Point Goal and Car Goal. The other environments use a fixed commanded goal in each episode.}
\label{fig: envs}
\end{figure}

We evaluate Safe-CRL on 12 failure-prone robot navigation and locomotion tasks built in Brax~\cite{freeman2021brax}, as illustrated in Figure~\ref{fig: envs}. Point Goal and Car Goal are adopted from the level-2 Safety-Gymnasium navigation tasks~\cite{ji2023safetygymnasium}, using the same environmental parameters. Entering an obstacle, hazard (pitfall), or moving-gremlin region causes failure, and the goal respawns after being reached. The remaining Ant and Humanoid tasks are adopted from Scaling-CRL~\cite{wang2025scalingcrl}. Falling is failure in Goal and Maze tasks, while Pitfall tasks additionally terminate when the robot enters a hazard region. Locomotion goals do not respawn after being reached. We use the MJX backend~\cite{todorov2012mujoco,zakka2025mujocoplayground} and make locomotion more failure-prone by reducing ground friction and, for Humanoid, reducing selected actuator gears. All methods share these modified dynamics. Following the lidar-style observations used in Safety-Gymnasium~\cite{ji2023safetygymnasium}, a 2D lidar is added to the observation to detect walls and obstacles. Occlusion is also considered in the lidar observation. Appendix~\ref{app:environments} provides detailed environment and failure specifications.

\paragraph{Baselines and implementations}
We use Scaling-CRL~\cite{wang2025scalingcrl} as the primary baseline because Safe-CRL directly extends its contrastive objective under failure termination, enabling a controlled comparison between Scaling-CRL and the proposed survival-mass correction. An additional hindsight-relabelling baseline, SAC-HER~\cite{haarnoja2018sac, andrychowicz2017her}, is reported in Appendix~\ref{app:sac_her}. For the main benchmark, both methods use 64-layer actor and critic networks, except for Point Goal and Car Goal, where 8-layer networks are used. Safe-CRL additionally uses a 4-layer Z-encoder, while all other training settings are identical to Scaling-CRL~\cite{wang2025scalingcrl}. We set $\gamma=0.99$ and the episode time limit to 1000 steps for all environments.

\paragraph{Evaluation metrics}
We report three complementary metrics: (1) \textit{time at goal}, the number of episode steps spent within the commanded goal region, jointly reflecting goal attainment and persistence near the goal~\cite{wang2025scalingcrl,bortkiewicz2025jaxgcrl}; (2) \textit{survival time}, the number of steps before failure termination or the episode time limit; and (3) \textit{goal coverage}, the percentage of episodes in which the commanded goal is reached by the robot at least once. Comparing time at goal and goal coverage indicates whether higher time at goal is obtained at the expense of reaching fewer commanded goals. Each metric is averaged over 128 parallel evaluation environments. We report the mean and standard deviation over five independent random seeds for the main benchmark and depth-scaling experiments, and over three seeds for the remaining experiments and ablations, following Scaling-CRL~\cite{wang2025scalingcrl}.

\section{Results and discussion}
\label{sec: results}

\paragraph{Main benchmark} 

\begin{figure}[h!]
\begin{center}
\includegraphics[width=0.85\columnwidth]{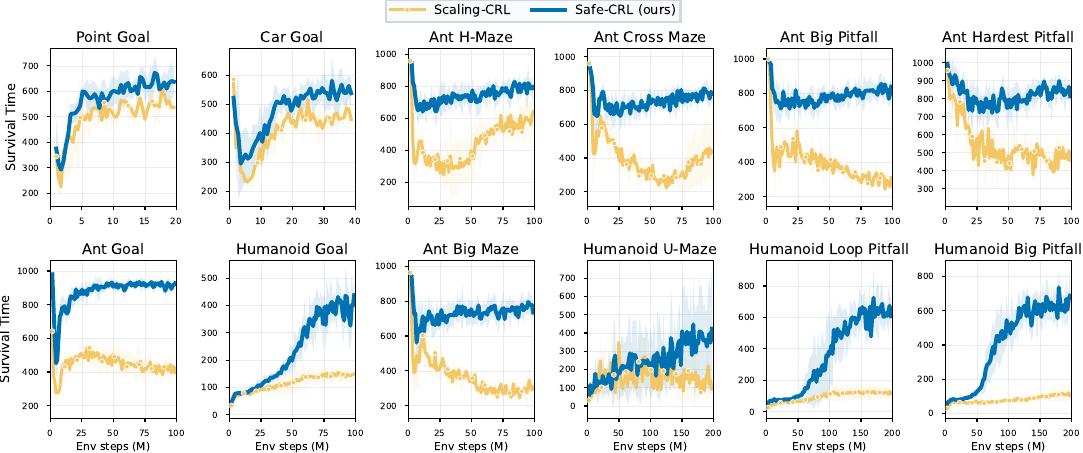}
\includegraphics[width=0.85\columnwidth]{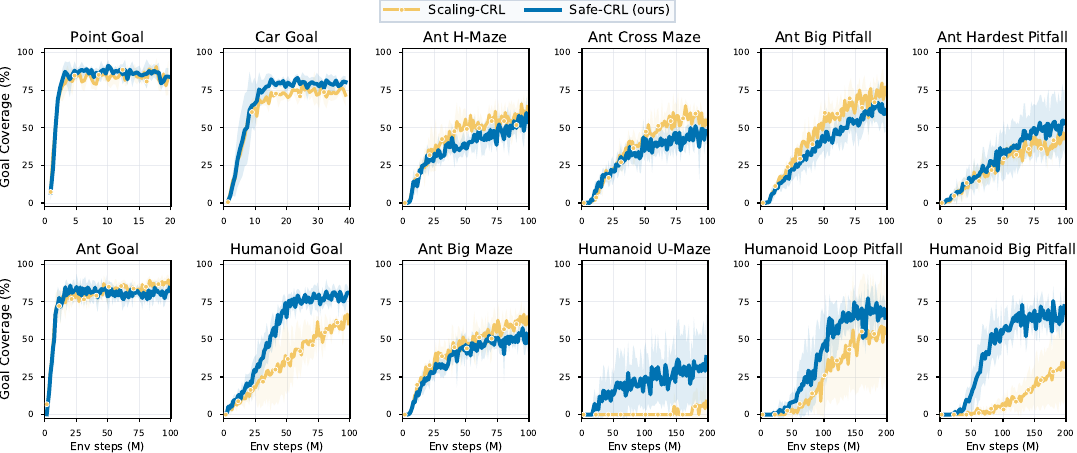}
\end{center}
\caption{Survival time (top two rows) and goal coverage (bottom two rows) on the 12-task benchmark (mean $\pm$ std).}
\label{fig: main_survival_coverage}
\end{figure}

Figure~\ref{fig: open-figure} shows time at goal, while Figure~\ref{fig: main_survival_coverage} compares survival time and goal coverage; Table~\ref{tab: main-number} provides the corresponding values. Safe-CRL achieves higher mean time at goal than Scaling-CRL in all 12 tasks. The gains are modest on Point Goal, Car Goal, and Ant H-Maze, but pronounced on the remaining nine tasks. For example, average time at goal increases from 276.8 to 684.0 steps on Ant Goal, from 22.2 to 266.4 on Humanoid Goal, and from 8.2 to 474.1 on Humanoid Big Pitfall. Safe-CRL also achieves higher mean survival time in all 12 tasks. Goal coverage increases in seven tasks and decreases in the remaining five. Despite these decreases, time at goal increases substantially in the affected environments, indicating that the gains are not simply obtained by sacrificing goal-reaching coverage.  

Comparing the three metrics provides further evidence for the two failure modes discussed in Section \ref{sec: intro}. In Ant Goal, Ant Big Maze, and Ant Hardest Pitfall, differences in goal coverage are much smaller than the corresponding differences in time at goal and survival time. This pattern indicates that Scaling-CRL can reach the commanded goal but often fails to remain there safely. This is \textit{unsustainable goal reaching}. For the more failure-prone Humanoid environments, Scaling-CRL's time at goal and goal coverage grow much more slowly than Safe-CRL, while its survival time remains low throughout training. This behaviour is consistent with \textit{catastrophic failure bootstrapping}, where repeated reinforcement of near-failure trajectories prevents the policy from escaping failure-prone behaviours. Overall, these learning patterns of Scaling-CRL are consistent with the two failure modes discussed in Section \ref{sec: intro}. Safe-CRL mitigates both failure modes and improves both goal-reaching performance and survival, with the largest gains in Humanoid tasks.

% These learning patterns are consistent with the two failure modes discussed in Section~\ref{sec: gap}: unstable goal reaching and catastrophic failure bootstrapping.

\begin{figure}[h!]
\begin{center}
%\framebox[4.0in]{$\;$}
\includegraphics[width=0.71\columnwidth]{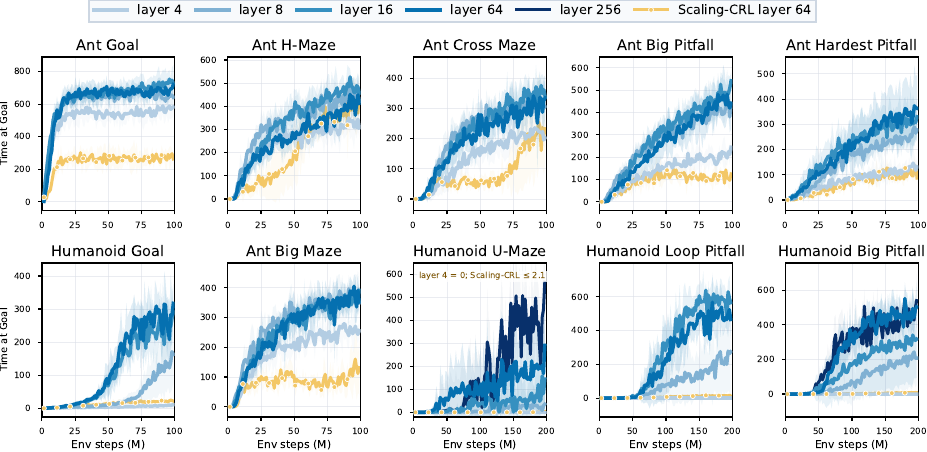}
\includegraphics[width=0.71\columnwidth]{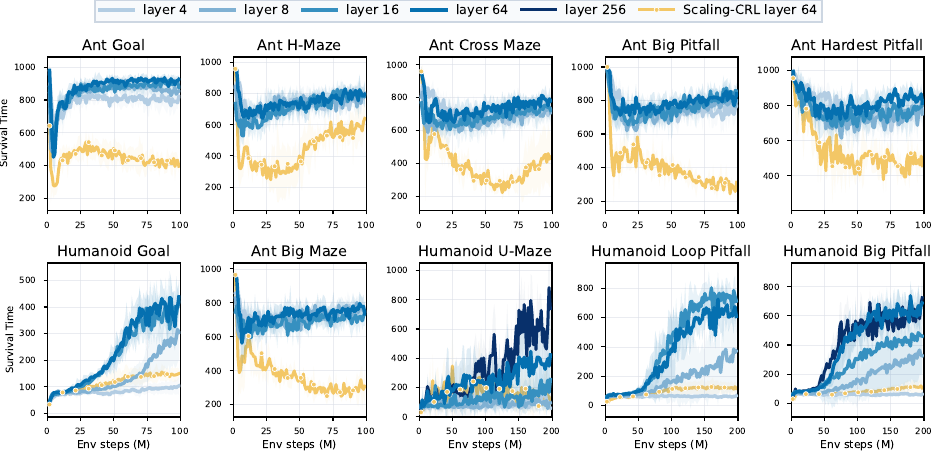}
\end{center}
\caption{Time at goal and survival time for Safe-CRL using different actor and critic depths. The Z-encoder depth is 4. Point Goal and Car Goal are omitted from the depth-scaling study because they are easier than locomotion.}
\label{fig: depth-scale-goal}
\end{figure}

\paragraph{Safe-CRL preserves depth scalability}
A key finding of Scaling-CRL is that goal-reaching performance improves with deeper networks~\cite{wang2025scalingcrl}. To examine whether Safe-CRL preserves this depth scalability, we fix the 4-layer Z-encoder and scale the actor and critic jointly from 4 to 64 layers, further extending to 256 layers for the more challenging Humanoid U-Maze and Humanoid Big Pitfall tasks. Figure~\ref{fig: depth-scale-goal} reports time at goal and survival time over five random seeds, with 64-layer Scaling-CRL included as a reference.

Figure~\ref{fig: depth-scale-goal} shows that time at goal and survival time generally improve as network depth increases. Their performance often saturates around 8--16 layers in the easier Ant tasks. Harder Humanoid tasks benefit from greater depth: the improvement on Humanoid Big Pitfall is pronounced up to approximately 64 layers, and Humanoid U-Maze continues to improve up to 256 layers. Notably, 8-layer Safe-CRL matches or outperforms 64-layer Scaling-CRL across all tasks evaluated in this depth-scaling study. These results show that Safe-CRL preserves CRL's depth scalability and further improves sample efficiency. 

\begin{figure}[h!]
\centering
    \includegraphics[width=0.55\columnwidth]{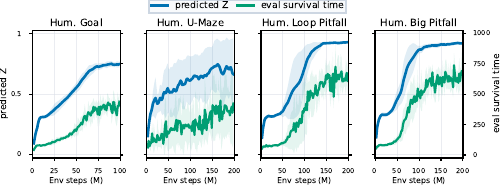}
    \caption{Predicted survival mass versus survival time in 128 evaluation runs. Actor and critic depths are 64.}
    \label{fig: pred_z_survival}
\end{figure}

\paragraph{Z-encoder learns meaningful survival mass with little overhead} 
We next examine whether the lightweight Z-encoder learns a meaningful survival-mass estimate. Figure~\ref{fig: pred_z_survival} compares predicted survival mass with observed survival time during training on the four Humanoid tasks. The quantities evolve with closely aligned trends: as predicted survival mass increases, evaluated survival time generally increases as well. The aligned trends indicate that the predicted survival mass tracks failure-dependent trajectory quality during training.

The Z-encoder is computationally lightweight. Using the default 4-layer architecture increases the total number of model parameters by only approximately 4.2\% (64-layer actor and critic networks) and the average wall-clock training time by 2.8\% (Table \ref{tab:training-time}). Thus, the corrections introduce only modest computational overhead.

\begin{figure}[h!]
\begin{center}
%\framebox[4.0in]{$\;$}
\includegraphics[width=0.9\columnwidth]{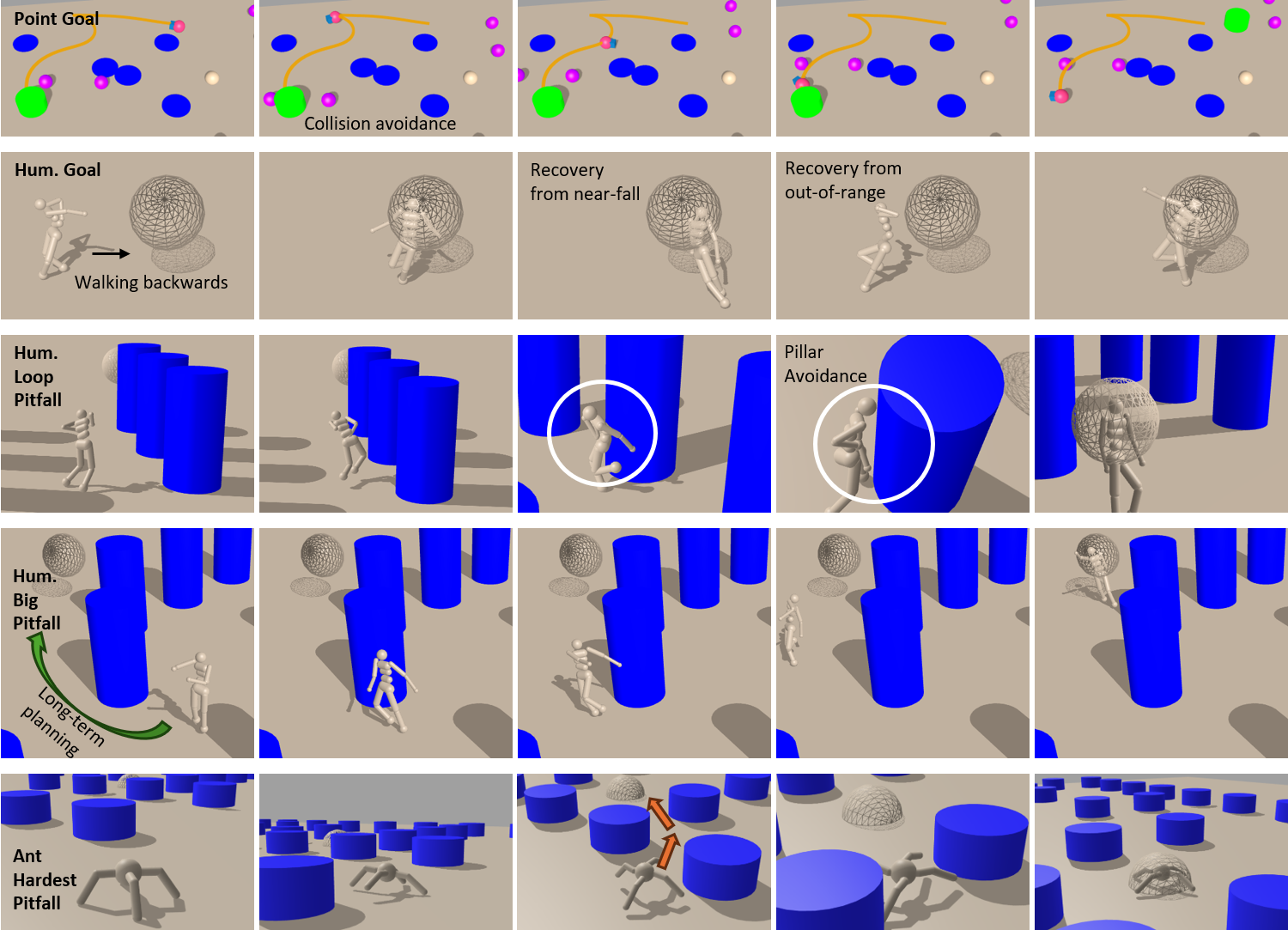}
\end{center}
\caption{Visualization of the policy learned by Safe-CRL. Each row contains 5 snapshots captured from one evaluation episode, showing the key failure-avoidance behaviours of the agent.}
\label{fig: vis_examples}
\end{figure}

\paragraph{Deeper Safe-CRL policies exhibit complex failure-avoidance behaviours}
We finally visualize representative rollouts of the learned 64-layer policies (8-layer for Point Goal and Car Goal) in Figure~\ref{fig: vis_examples}. Despite receiving only a one-bit failure signal, Safe-CRL learns diverse failure-avoidance behaviours while reaching the commanded goal. On Point Goal, the robot brakes and replans an S-shaped trajectory to pass between moving gremlins after the goal respawns. On Humanoid Goal, the robot walks backward toward a goal behind it and recovers from near-falls and overshooting after reaching the goal. On Humanoid Loop Pitfall, the humanoid adjusts its body orientation to avoid nearby hazard regions, while on the more difficult pitfall tasks, the humanoid and ant exhibit longer-horizon behaviours such as taking detours or slowing down to pass safely between hazards. These rollouts illustrate non-trivial failure-avoidance and recovery behaviours learned from a sparse one-bit failure signal. Additional rollout visualizations are available on the project page \hyperlink{https://github.com/RomainLITUD/safe-crl}{https://github.com/RomainLITUD/safe-crl}.

Additional experimental results are provided in Appendix \ref{app:supplementary}.

\section{Ablations}

In this section, we mainly ablate the two corrections in Safe-CRL and examine the sensitivity to the coefficient of the $\log Z$ term. Additional ablations on Z-encoder depth, TD-based survival-mass estimation, and the effect of goal respawning are provided in Appendix~\ref{app:supplementary}.

\paragraph{Core component ablations}
Table~\ref{tab: main-ablation} isolates the contributions of mass-weighted InfoNCE and the log-survival-mass correction. MW-InfoNCE alone produces only modest changes relative to Scaling-CRL, whereas the log-$Z$ correction accounts for the dominant improvement across the four tasks. For example, on Humanoid Goal, the log-$Z$-only variant increases time at goal from 13.8 to 288.7 steps and survival time from 113.9 to 397.0 steps, while mass-weighted InfoNCE alone achieves 28.3 and 146.1 steps, respectively. A similar pattern is observed on Ant Big Maze and Ant Hardest Pitfall. This observation is reasonable because the $\log{Z}$ term influences the policy learning directly, while the critic-side MW-InfoNCE corrects the distortion in contrastive learning but does not restore the missing survival mass for policy update, as explained in Section \ref{sec: bias role}.

\begin{table*}[h!]
\centering
\caption{Core component ablation of Safe-CRL. MW denotes MW-InfoNCE. Results are averaged over the final 10\% of training and reported as mean $\pm$ standard deviation over three random seeds. The actor and critic depths are 16 (8 for Car Goal), and the Z-encoder depth is 4.}
\label{tab:component_ablation}
\resizebox{\textwidth}{!}{
\begin{tabular}{lcc|cc|cc|cc|cc}
\toprule
& \multicolumn{2}{c|}{}
& \multicolumn{2}{c|}{Car Goal}
& \multicolumn{2}{c|}{Humanoid Goal}
& \multicolumn{2}{c|}{Ant Big Maze}
& \multicolumn{2}{c}{Ant Hardest Pitfall} \\
Method
& MW & $\log Z$
& Time at goal & Survival time
& Time at goal & Survival time
& Time at goal & Survival time
& Time at goal & Survival time \\
\midrule
Scaling-CRL
& \xmark & \xmark
& 2.2$\pm$0.1 & 464.9$\pm$11.2 & 13.8$\pm$1.0 & 113.9$\pm$5.3 & 89.7$\pm$34.2 & 261.1$\pm$33.1 & 98.5$\pm$17.1 & 381.1$\pm$11.6 \\

MW only
& \cmark & \xmark
& 2.2$\pm$0.1 & 471.4$\pm$13.3 & 28.3$\pm$0.7 & 146.1$\pm$10.2 & 91.4$\pm$31.1 & 291.6$\pm$51.6 & 113.2$\pm$19.3 & 411.0$\pm$27.1 \\

$\log Z$ only
& \xmark & \cmark
& \textbf{2.4}$\pm$0.1 & 537.1$\pm$61.3 & 288.7$\pm$21.6 & 397.0$\pm$21.8 & 337.0$\pm$28.9 & 693.5$\pm$33.7 & 271.2$\pm$28.3 & 722.9$\pm$32.6 \\

Safe-CRL
& \cmark & \cmark
& \textbf{2.4}$\pm$0.2 & \textbf{555.5}$\pm$49.1 & \textbf{297.0}$\pm$19.4 & \textbf{413.4}$\pm$23.9 & \textbf{352.9}$\pm$11.1 & \textbf{725.7}$\pm$31.7 & \textbf{313.0}$\pm$50.7 & \textbf{787.9}$\pm$51.1 \\
\bottomrule
\end{tabular}}
\label{tab: main-ablation}
\end{table*}

\paragraph{Ablations on $\log Z$'s weight}

\begin{figure}[h!]
\begin{center}
%\framebox[4.0in]{$\;$}
\includegraphics[width=0.49\columnwidth]{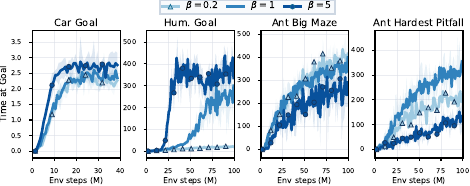}
\includegraphics[width=0.49\columnwidth]{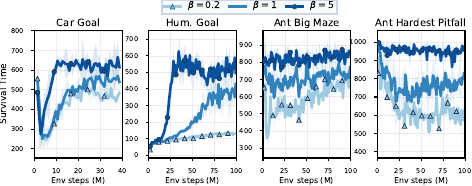}
\end{center}
\caption{Ablations on the coefficient $\beta$ of $\log Z$. The left four panels report time at goal, and the right four panels report survival time. We set the actor and critic depths to 16 layers (8 for Car Goal) and vary $\beta$.}
\label{fig: log-Z coefficient}
\end{figure}

Under the score convention used in Section~\ref{sec: methodology}, the failure-aware occupancy derivation fixes the coefficient of $\log Z$ to one. To examine deviations from this value, we introduce an analysis coefficient $\beta$ and compare 0.2, 1.0, and 5.0 in Figure~\ref{fig: log-Z coefficient}. Increasing $\beta$ generally increases survival time, whereas its effect on goal-reaching performance is environment-dependent. Safe-CRL fixes $\beta=1$ because it is theory-derived rather than tuned as a safety--goal trade-off parameter; the experiment does not assume that this value maximizes every empirical metric in every environment.

\section{Conclusion}

This paper studied contrastive reinforcement learning under failure termination and identified a systematic overestimation bias caused by the missing survival mass of the failure-aware discounted future-goal occupancy. The analysis of this bias led to two minimal corrections, mass-weighted InfoNCE for critic learning and a log-survival-mass correction for policy optimization, which together form Safe-CRL. Across failure-prone navigation and locomotion tasks, Safe-CRL improves survival and goal-reaching performance while preserving the depth scalability of CRL, using only the one-bit failure signal provided by termination. More broadly, our results suggest that under failure termination, scalable self-supervised policy learning should preserve not only the relative structure of observed futures, but also the total occupancy mass removed by termination.

\paragraph{Limitations}
We point out three limitations here: First, Safe-CRL is designed for failure-terminated goal-conditioned control and does not directly address non-terminating safety costs or constrained-RL objectives. Second, time-limit truncation also introduces right censoring. In this study, we retain CRL’s finite-support treatment for future relabelling and treat Z-encoder horizons without an observed failure as survived. Future work could incorporate the right-censoring correction proposed in SVL~\cite{tiofack2026svl}. Finally, Safe-CRL retains the exploration limitations of CRL, which can result in uneven goal-space coverage in difficult tasks (see an analysis in Figure~\ref{fig: limit-on-explore}, Appendix~\ref{app:supplementary}). Addressing these settings is also left for future work.

\section{Related work}

Finally, we briefly review the work most closely related to this study. Goal-conditioned reinforcement learning (GCRL) commonly uses achieved future states through hindsight relabelling, transforming sparse goal-reaching experience into dense self-supervision~\cite{andrychowicz2017her}. C-learning~\cite{eysenbach2021clearning} formulates goal-conditioned value estimation as classification between future and independently sampled states, while contrastive reinforcement learning (CRL)~\cite{eysenbach2022contrastiverl} connects contrastive density-ratio estimation to discounted future-state occupancy and goal-conditioned value learning. More recently, Scaling-CRL~\cite{wang2025scalingcrl} showed that this self-supervised objective can exploit substantially deeper networks, establishing CRL as a promising regime for scalable self-supervised RL.

For CRL specifically, several studies have examined whether distributions constructed from observed future outcomes preserve all information required for goal-conditioned control. Distributional Distance Classifiers (DDC)~\cite{akella2023distributional} separate horizon-dependent goal-reaching quantities from reachability probability. USHER~\cite{schramm2023usher} corrects the Hindsight-selection bias under stochastic dynamics, where conditioning on observed outcomes can under-represent unfavourable transitions and bias goal-conditioned estimates. These works show that conditioning or learning from observed future outcomes can distort goal-reaching quantities in ways that matter for control. Their focus is horizon dependence or stochastic hindsight selection, whereas our work studies a different methodological issue: the loss of total discounted future-goal occupancy mass when failure-terminated pre-failure futures are normalized. Recent work also extends single-goal contrastive RL to safety-aware exploration by exploiting its representation-induced exploration mechanism~\cite{bastankhah2026demystifying}.

A complementary line of recent studies reformulates GCRL through survival analysis. Survival Value Learning (SVL)~\cite{tiofack2026svl} models time to goal using event and right-censored trajectories, while Survival Reinforcement Learning (SRL)~\cite{tiofack2026srl} extends this formulation to online self-supervised learning and introduces goal ``dwell-time'' to encourage stable long-horizon behaviour. These works are conceptually adjacent in highlighting temporal information overlooked by standard goal-reaching objectives, but address a different formulation: SVL and SRL explicitly model time-to-event distributions, whereas Safe-CRL retains the CRL occupancy formulation and identifies a normalization-induced overestimation bias caused by missing survival mass under failure termination.

It is useful to mention a separate line of safe reinforcement learning that formulates safety through constrained Markov decision processes (CMDPs), where a policy maximizes task reward subject to explicit constraints on expected cumulative costs~\cite{altman1999constrained}. Representative methods include Constrained Policy Optimization (CPO), which directly enforces policy constraints during optimization~\cite{achiam2017constrained}, Lyapunov-based safe RL~\cite{chow2018lyapunov}, and Lagrangian methods~\cite{stooke2020responsive}. These methods address an important but different safety setting from ours: they assume an explicit cost or constraint signal and optimize a constrained objective. An ``unsafe event'' does NOT necessarily terminate the episode. Safe-CRL instead considers failure-terminated goal-conditioned learning, where termination provides the only one-bit failure signal.

In summary, existing work has studied scalable CRL, stochastic and horizon-dependent goal reachability, and survival-based goal-conditioned learning. To our knowledge, prior work has not identified the systematic overestimation bias that arises when established CRL normalizes pre-failure futures and discards the survival mass intrinsic to the failure-aware discounted future-goal occupancy, nor the accompanying trajectory-wise normalization distortion in critic learning.

%Bibliography
\bibliographystyle{unsrt}  
\bibliography{references}

\newpage
\appendix

\counterwithin{figure}{section}
\counterwithin{table}{section}
\numberwithin{equation}{section}

\section{Proofs and technical details}
\label{app: proof}

This appendix provides the detailed derivations for the theoretical results in Sections \ref{sec: gap} and \ref{sec: methodology}. We first show that the failure-aware discounted future-goal occupancy is the correct goal-conditioned value, then prove that survival-mass weighting in the InfoNCE loss corrects the trajectory-wise normalization distortion induced by future relabelling. Finally, we derive the log-survival-mass correction for actor optimization.

\subsection{Proof of Proposition 1}
\label{app:proof_prop1}

We prove Proposition 1 using measurable goal sets, which avoids point-mass notation for continuous goal spaces. For an anchor $x=(s,a)$ and a measurable goal set $B\subseteq\mathcal{G}$, define the failure horizon as
\begin{equation}
T_f \triangleq \inf \left\{h\geq 1:
F(s_{t+h-1},a_{t+h-1},s_{t+h})=1 \right\},
\end{equation}
with $T_f=\infty$ if no failure occurs. Therefore, $T_f>h$ means that $S_{t+h}$ lies on the valid pre-failure portion of the trajectory. The corresponding failure-aware discounted future-goal occupancy is
\begin{equation}
\mu_f^\pi(B\mid x) = \sum_{h=1}^{\infty}q_\gamma(h)
\Pr^\pi\left(\phi(S_{t+h})\in B, T_f>h \mid x_t=x \right),
\quad
q_\gamma(h)=(1-\gamma)\gamma^{h-1}.
\end{equation}

Consider the normalized goal-reaching transition reward
\begin{equation}
r_B(s,a,s') =  (1-\gamma) \mathbf{1}\left\{ F(s,a,s')=0, \phi(s')\in B \right\}.
\end{equation}
The corresponding discounted action value is
\begin{equation}
Q_B^\pi(s,a) = \mathbb{E}^{\pi}\left[
\sum_{k=0}^{\infty}
\gamma^k
r_B(s_{t+k},a_{t+k},s_{t+k+1}) \mid s_t=s,a_t=a\right].
\end{equation}
Because failure immediately terminates the episode, a valid goal-reaching reward at horizon $k+1$ is obtained only when $T_f>k+1$. Substituting the reward definition gives
\begin{equation}
Q_B^\pi(s,a) =
\sum_{k=0}^{\infty} (1-\gamma)\gamma^k \Pr^\pi\left( \phi(s_{t+k+1})\in B, T_f>k+1\mid x_t=x \right).
\end{equation}
Letting $h=k+1$ directly yields
\begin{equation}
Q_B^\pi(s,a) = 
\sum_{h=1}^{\infty} q_\gamma(h)\Pr^\pi\left(\phi(S_{t+h})\in B, T_f>h\mid x_t=x \right)
=
\mu_f^\pi(B\mid x).
\end{equation}
This proves Proposition 1. For discrete goals, or equivalently using density notation for continuous goal spaces, this gives the pointwise relation used in the main text,
\begin{equation}
Q_g^\pi(s,a)=\mu_f^\pi(g\mid x).
\end{equation}
Taking $B=\mathcal{G}$ further gives
\begin{equation}
Z^\pi(x) = \mu_f^\pi(\mathcal{G}\mid x)
= \sum_{h=1}^{\infty} q_\gamma(h) \Pr^\pi(T_f>h\mid x),
\end{equation}
which is exactly the defined survival mass.

\subsection{Proof of Proposition 2}
\label{app:proof_prop2}

Consider an anchor $x_t=x=(s,a)$ and a realized trajectory $\tau$ from this anchor. Let $L_\tau$ denote the number of valid future states before failure. For a measurable goal set $B\subseteq\mathcal{G}$, define the realized failure-aware discounted future-goal measure as
\begin{equation}
\hat{\mu}_\tau(B\mid x)
=
\sum_{h=1}^{L_\tau} q_\gamma(h) \mathbf{1}\{\phi(S_{t+h})\in B\}.
\end{equation}
Its total mass is
\begin{equation}
\hat Z_\tau = \hat{\mu}_\tau(\mathcal{G}\mid x)
= \sum_{h=1}^{L_\tau} q_\gamma(h)
= 1-\gamma^{L_\tau}.
\end{equation}
Conditioned on the realized trajectory $\tau$, established CRL samples positive goals only from these valid pre-failure futures. Therefore, for $\hat Z_\tau>0$, the corresponding normalized positive-goal distribution is
\begin{equation}
r_\tau(B\mid x) = \frac{\hat{\mu}_\tau(B\mid x)}{\hat Z_\tau}.
\end{equation}
Without mass weighting, averaging this distribution over trajectories gives the established CRL positive distribution,
\begin{equation}
p_{\mathrm{CRL}}^+(B\mid x)
= \mathbb{E}_{\tau\mid x} \left[ r_\tau(B\mid x) \right]
= \mathbb{E}_{\tau\mid x} \left[ \frac{\hat{\mu}_\tau(B\mid x)}{\hat Z_\tau} \right].
\end{equation}
In general,
\begin{equation}
\mathbb{E}_{\tau\mid x}
\left[\frac{\hat{\mu}_\tau(B\mid x)}{\hat Z_\tau}\right]
\neq \frac{\mathbb{E}_{\tau\mid x}\left[\hat{\mu}_\tau(B\mid x)\right]}{\mathbb{E}_{\tau\mid x}
\left[\hat Z_\tau\right]
}.
\end{equation}
Thus, trajectory-wise normalization does not generally recover the normalized population failure-aware occupancy. We now consider the mass-weighted InfoNCE objective. Multiplying the complete contrastive row associated with trajectory $\tau$ by $\hat Z_\tau$ induces the weighted positive measure
\begin{equation}
\nu_W(B\mid x)
\triangleq \mathbb{E}_{\tau\mid x} \left[ \hat Z_\tau r_\tau(B\mid x) \right].
\end{equation}
Using the definition of $r_\tau$, the trajectory-wise normalization cancels exactly:
\begin{equation}
\hat Z_\tau r_\tau(B\mid x) = \hat{\mu}_\tau(B\mid x).
\end{equation}
Therefore,
\begin{equation}
\nu_W(B\mid x)
=\mathbb{E}_{\tau\mid x}\left[\hat{\mu}_\tau(B\mid x)\right]
=\mu_f^\pi(B\mid x).
\end{equation}
Taking $B=\mathcal{G}$ gives the total weighted mass
\begin{equation}
\nu_W(\mathcal{G}\mid x)
=\mathbb{E}_{\tau\mid x}\left[\hat Z_\tau\right]
= Z^\pi(x).
\end{equation}
So, the normalized positive distribution induced by mass-weighted InfoNCE is
\begin{equation}
p_W^+(B\mid x)
=\frac{\nu_W(B\mid x)}{\nu_W(\mathcal{G}\mid x)}
=\frac{\mu_f^\pi(B\mid x)}{Z^\pi(x)}.
\end{equation}
Equivalently, using the density notation adopted in the main text,
\begin{equation}
p_W^+(g\mid x)
= \frac{\mu_f^\pi(g\mid x)}{Z^\pi(x)}
= \bar{\mu}_f^\pi(g\mid x).
\end{equation}
This proves Proposition 2. Mass weighting therefore corrects the trajectory-wise normalization distortion and recovers the correct conditional shape of the failure-aware occupancy. However,
\begin{equation}
\int_{\mathcal{G}}p_W^+(g\mid x)\,dg=1,
\end{equation}
so the contrastive distribution remains normalized, and the total survival mass $Z^\pi(x)$ is still absent from the critic score. This remaining factor is restored in actor optimization by the log-survival-mass correction derived in Appendix \ref{app:proof_corollary}.

For anchors with $Z^\pi(x)=0$, the normalized distribution above is undefined, but such anchors receive zero weight in the mass-weighted population objective; this case is also discussed separately in Appendix \ref{app:technical_conditions}.

\paragraph{Finite-minibatch estimator}

For a mini-batch of $B$ independently sampled anchor-trajectory pairs, the implemented critic loss is
\begin{equation}
\widehat{\mathcal{L}}_{\mathrm{MW\text{-}NCE}}
= \frac{1}{B}\sum_{i=1}^{B}\hat Z_i\ell_i^{\mathrm{NCE}},
\end{equation}
where $\ell_i^{\mathrm{NCE}}$ denotes the complete row-wise InfoNCE loss for anchor $i$. Its expectation satisfies
\begin{equation}
\mathbb{E}\left[\widehat{\mathcal{L}}_{\mathrm{MW\text{-}NCE}} \right]
=\mathbb{E}\left[\hat Z\ell^{\mathrm{NCE}} \right].
\end{equation}
Thus, the unnormalized mini-batch average is a direct Monte Carlo estimator of the population mass-weighted contrastive risk. No normalization of the weights within each mini-batch is required.

In particular, the self-normalized alternative
\begin{equation}
\frac{\sum_{i=1}^{B}\hat Z_i\ell_i^{\mathrm{NCE}}}{\sum_{i=1}^{B}\hat Z_i}
\end{equation}
contains a random denominator and is not the same finite-minibatch estimator as the objective used in Safe-CRL.

The negative goals appearing in each InfoNCE row remain sampled from the ordinary mini-batch goal marginal, as in established CRL. Row weighting changes the contribution of the anchor-positive pair to the population risk but does not require the negative proposal to be re-weighted. The resulting negative proposal therefore remains anchor-independent, as required by the density-ratio interpretation used in Appendix \ref{app:proof_corollary}.

\subsection{Proof of Corollary 1}
\label{app:proof_corollary}

Let $p^-(g)$ denote the anchor-independent negative proposal. Under the support condition stated in Appendix A.5, the population optimum of InfoNCE identifies the positive-to-negative density ratio up to an anchor-dependent additive offset:
\begin{equation}
f_W^*(x,g)
=\log\frac{p_W^+(g\mid x)}{p^-(g)}+c(x).
\end{equation}
Because $x=(s,a)$, the offset $c(x)$ may in general depend on the action. Therefore, the row-wise InfoNCE objective alone does not make scores directly comparable across different actions. To remove this arbitrary offset, define the calibrated contrastive score
\begin{equation}
\bar f_W(x,g)
=f_W(x,g)-\log\mathbb{E}_{g'\sim p^-}\left[\exp f_W(x,g')\right].
\end{equation}
At the population optimum,
\begin{equation}
\mathbb{E}_{g'\sim p^-}\left[\exp f_W^*(x,g')\right]
=\mathbb{E}_{g'\sim p^-}\left[\exp\left(\log\frac{p_W^+(g'\mid x)}{p^-(g')}+c(x)\right)
\right].
\end{equation}
Therefore,
\begin{equation}
\mathbb{E}_{g'\sim p^-}\left[\exp f_W^*(x,g')\right]
=e^{c(x)}\int_{\mathcal{G}}p^-(g')\frac{p_W^+(g'\mid x)}{p^-(g')}\,dg'.
\end{equation}

Because $p_W^+(\cdot\mid x)$ is a normalized distribution,
\begin{equation}
\int_{\mathcal{G}}p_W^+(g'\mid x)\,dg'
=1,
\end{equation}
and therefore
\begin{equation}
\mathbb{E}_{g'\sim p^-}\left[\exp f_W^*(x,g')\right]= e^{c(x)}.
\end{equation}
Substituting this result into the calibrated score gives
\begin{equation}
\bar f_W^*(x,g)=\log\frac{p_W^+(g\mid x)}{p^-(g)}.
\end{equation}

Using Proposition 2, we have:
\begin{equation}
p_W^+(g\mid x)=\frac{\mu_f^\pi(g\mid x)}{Z^\pi(x)},
\end{equation}
so that
\begin{equation}
\bar f_W^*(x,g)=\log\frac{\mu_f^\pi(g\mid x)}{Z^\pi(x)p^-(g)}.
\end{equation}
Equivalently,
\begin{equation}
\bar f_W^*(x,g)=\log \mu_f^\pi(g\mid x)-\log Z^\pi(x)-\log p^-(g).
\end{equation}
Adding the log survival mass therefore yields
\begin{equation}
\bar f_W^*(x,g)+\log Z^\pi(x)
=\log \mu_f^\pi(g\mid x)-\log p^-(g).
\end{equation}

For a fixed state $s$ and commanded goal $g$, the negative proposal $p^-(g)$ is independent of the candidate action $a$. So, we have:
\begin{equation}
\arg\max_a\left[\bar f_W^*(s,a,g)+\log Z^\pi(x)\right]=\arg\max_a\log\mu_f^\pi(g\mid x).
\end{equation}
Because the logarithm is strictly increasing,
\begin{equation}
\arg\max_a\left[\bar f_W^*(s,a,g)+\log Z^\pi(x)\right]
=\arg\max_a\mu_f^\pi(g\mid x).
\end{equation}
Finally, Proposition 1 gives
\begin{equation}
\mu_f^\pi(g\mid x)=Q_g^\pi(s,a),
\end{equation}
and
\begin{equation}
\arg\max_a\left[\bar f_W^*(s,a,g)+\log Z^\pi(x)\right]
=\arg\max_a Q_g^\pi(s,a).
\end{equation}

This proves Corollary 1. The coefficient of the log-$Z$ correction is exactly 1.0 because the failure-aware occupancy factorizes multiplicatively as
\begin{equation}
\mu_f^\pi(g\mid x)=Z^\pi(x)p_W^+(g\mid x).
\end{equation}

So, $\log Z^\pi(x)$ is not introduced as a trade-off term, but follows directly from restoring the total mass removed by contrastive normalization.

\subsection{Population target of the Z-encoder}
\label{app:z_population}

The Z-encoder is trained from a one-sample Monte Carlo survival target. For a trajectory $\tau$, we sample $H\sim q_\gamma$ and define $Y=\mathbf{1}\{T_f>H\}$. For the uncensored process,
\begin{equation}
\mathbb{E}_{H}[Y\mid\tau]
=
\sum_{h=1}^{L_\tau} q_\gamma(h)
=
\hat Z_\tau,
\qquad
\mathbb{E}_{\tau,H}[Y\mid x]
=
\mathbb{E}_{\tau}[\hat Z_\tau\mid x]
=
Z^\pi(x).
\end{equation}
For a fixed anchor $x$, the corresponding binary cross-entropy risk is
\begin{equation}
\mathcal{L}_x(z)
=
-\mathbb{E}
\left[
Y\log z+(1-Y)\log(1-z)
\mid x
\right],
\qquad z\in(0,1).
\end{equation}
Its population minimizer is
\begin{equation}
Z_\psi^\ast(x)
=
\mathbb{E}[Y\mid x]
=Z^\pi(x).
\end{equation}

Because binary cross-entropy is affine in its target label, sampling one $Y$ provides an unbiased Monte Carlo estimator of the soft-target BCE based on $\hat Z_\tau$. Under time-limit truncation, horizons without an observed failure are treated as survived, yielding the finite-support approximation discussed in Appendix~\ref{app:technical_conditions}.

\subsection{Technical conditions}
\label{app:technical_conditions}

The theoretical results above rely on the following conditions.

\paragraph{(1) Replay-induced trajectory distribution}
The notation $\pi$ denotes the replay-induced trajectory distribution represented by the data used for contrastive learning. When replay contains trajectories collected under different commanded goals, $\pi$ is the corresponding goal-marginalized replay-induced trajectory distribution; the same replay-induced trajectory distribution must be used consistently in the definitions of $\mu_f^\pi$ and $Z^\pi$.

\paragraph{(2) Support of the negative proposal}
For the contrastive density ratio to be well-defined, the negative proposal satisfies
\begin{equation}
p^-(g)>0\qquad\text{whenever}\qquad p_{\mathrm W}^+(g|x)>0.
\end{equation}
This is the standard support condition for contrastive density-ratio estimation.

\paragraph{(3) Continuous goal spaces}
For continuous goal spaces, expressions such as $\mu_f^\pi(g|x)$ and $p_{\mathrm W}^+(g|x)$ denote densities with respect to a common dominating measure. Equivalently, all results can be stated directly for measurable goal sets without assuming the existence of densities.

\paragraph{(4) Zero-mass trajectories}
If a trajectory contains no valid future state for an anchor, then
\begin{equation}
L_\tau=0, \qquad\hat Z_\tau=0.
\end{equation}

Such an anchor has zero weight in the mass-weighted contrastive objective and therefore does not require a valid positive goal in the population formulation. In implementation, rows with $L_\tau=0$ are masked from the InfoNCE objective before positive-goal sampling.

\paragraph{(5) Failure termination versus right censoring}
The theoretical quantities $\mu_f^\pi$ and $Z^\pi$ are defined for the uncensored failure-terminated process. In implementation, following Scaling-CRL~\cite{wang2025scalingcrl,bortkiewicz2025jaxgcrl}, futures beyond the available replay support or time limit are marked as invalid, and we do not apply an explicit right-censoring correction. When this support ends because of time-limit truncation rather than failure, the trajectory is right-censored in survival-analysis terminology~\cite{tiofack2026svl}. Consequently, $\hat Z_t$ should be interpreted operationally as the realized survival mass of the future support available to the sampler. Importantly, the sample-wise identity underlying Proposition 2 remains exact on this observed support,
\begin{equation}
\hat Z_\tau r_\tau(g\mid x)=\hat\mu_\tau(g\mid x).
\end{equation}

Therefore, mass weighting still exactly cancels the trajectory-wise normalization introduced by the implemented future sampler. For Z-encoder training, we instead sample $H\sim q_\gamma$ and use $Y=\mathbf{1}\{T_f>H\}$. Horizons extending beyond the available trajectory without an observed failure are treated as valid survival samples with $Y=1$; horizons crossing an observed failure remain valid with $Y = 0$. Without censoring, $\mathbb{E}[Y\mid x]=Z^\pi(x)$. Under time-limit truncation, treating horizons without an observed failure as survived yields a finite-support approximation that can overestimate the uncensored $Z^\pi(x)$. We do not apply an explicit censoring correction.

\section{Environments}
\label{app:environments}

We evaluate Safe-CRL on 12 failure-prone goal-conditioned robot control tasks implemented in Brax~\cite{freeman2021brax} with the MJX backend~\cite{todorov2012mujoco,zakka2025mujocoplayground}, providing richer friction and contact models. In all environments, failure produces a one-bit failure signal and immediately terminates the episode. Time-limit truncation is not treated as failure.

\paragraph{Navigation tasks}
Point Goal and Car Goal are adopted from the level-2 Safety-Gymnasium navigation tasks~\cite{ji2023safetygymnasium}. Goals respawn after being reached, and entering a designated obstacle, hazard, or moving-gremlin region triggers failure termination. The number of obstacles, hazards, gremlins, and their sizes, layout generation rules, playground sizes, 3-channel 16-bin 2D lidar observations, and ego-robot states are all identical to the original Safety-Gymnasium package.

\paragraph{Locomotion tasks}
The Ant and Humanoid tasks are adapted from Scaling-CRL~\cite{wang2025scalingcrl} and JaxGCRL~\cite{bortkiewicz2025jaxgcrl}. Goals remain fixed after being reached. Falling terminates all locomotion tasks, while Pitfall variants additionally terminate when the robot enters a designated hazard region. Maze-wall contact itself is not treated as failure. For locomotion tasks, we also equip the agents with a 16-bin, one-channel 2D lidar to detect surrounding walls and pitfalls.

Table~\ref{tab:environments} summarizes the task-specific termination rules. To make the locomotion tasks sufficiently failure-prone, we reduce the ground sliding friction coefficient from 1.0 to 0.6, creating a slippery ground. In particular, for the Humanoid robot, the actuator gears of the abdomen, lower limbs, and arms are reduced, as listed in Table \ref{tab:gear}. These modified dynamics are shared by both the Scaling-CRL baseline and the proposed Safe-CRL.

\begin{table}[h!]
\centering
\caption{Summary of the experimental environments and failure mechanisms.}
\label{tab:environments}
\begin{tabular}{lccc}
\toprule
Environment & Agent & Goal Respawn & Failure Termination \\
\midrule
Point Goal              & Point    & Yes & Enter obstacle, hazard, or gremlin region \\
Car Goal                & Car      & Yes & Enter obstacle, hazard, or gremlin region \\
Ant Goal                & Ant      & No  & Fall \\
Humanoid Goal           & Humanoid & No  & Fall \\
Ant H-Maze              & Ant      & No  & Fall \\
Ant Cross Maze          & Ant      & No  & Fall \\
Ant Big Maze            & Ant      & No  & Fall \\
Humanoid U-Maze         & Humanoid & No  & Fall \\
Ant Big Pitfall         & Ant      & No  & Fall or enter hazard region  \\
Ant Hardest Pitfall     & Ant      & No  & Fall or enter hazard region  \\
Humanoid Loop Pitfall   & Humanoid & No  & Fall or enter hazard region \\
Humanoid Big Pitfall    & Humanoid & No  & Fall or enter hazard region \\
\bottomrule
\end{tabular}
\end{table}

\begin{table}[h!]
\centering
\caption{Humanoid actuator gears under the default setting and our experiments.}
\label{tab:humanoid-actuator-gears}
\small
\setlength{\tabcolsep}{4pt}
\begin{tabular}{lrrr}
\toprule
Actuator gear & default & ours\\
\midrule
\texttt{abdomen\_y}        & 350 & 100 \\
\texttt{abdomen\_z}        & 350 & 100\\
\texttt{abdomen\_x}        & 350 & 100\\
\texttt{right\_hip\_x}     & 350 & 100\\
\texttt{right\_hip\_z}     & 350 & 100\\
\texttt{right\_hip\_y}     & 350 & 300\\
\texttt{right\_knee}       & 350 & 200\\
\texttt{left\_hip\_x}      & 350 & 100\\
\texttt{left\_hip\_z}      & 350 & 100\\
\texttt{left\_hip\_y}      & 350 & 300\\
\texttt{left\_knee}        & 350 & 200\\
\texttt{right\_shoulder1}  & 100 &  25\\
\texttt{right\_shoulder2}  & 100 &  25\\
\texttt{right\_elbow}      & 100 &  25\\
\texttt{left\_shoulder1}   & 100 &  25\\
\texttt{left\_shoulder2}   & 100 &  25\\
\texttt{left\_elbow}       & 100 &  25\\
\bottomrule
\label{tab:gear}
\end{tabular}
\end{table}

\section{SAC-HER baseline}
\label{app:sac_her}

\begin{figure}[h!]
\begin{center}
%\framebox[4.0in]{$\;$}
\includegraphics[width=0.49\columnwidth]{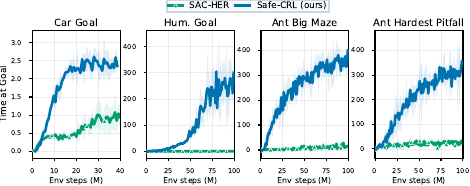}
\includegraphics[width=0.49\columnwidth]{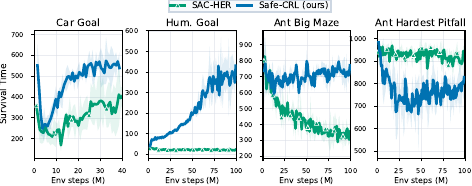}
\end{center}
\caption{Comparison with SAC-HER on four representative failure-prone tasks. Curves report time at goal and survival time. Safe-CRL uses 16 layers for the actor and critic networks (8 for Car Goal) and 4 layers for the Z-encoder.}
\label{fig:sac_her}
\end{figure}

We additionally compare Safe-CRL with SAC-HER, combining Soft Actor-Critic~\cite{haarnoja2018sac} with Hindsight Experience Replay~\cite{andrychowicz2017her}. The implementation is based on JaxGCRL~\cite{bortkiewicz2025jaxgcrl}. Scaling-CRL has previously shown that contrastive self-supervision substantially outperforms conventional HER-based goal-conditioned RL in large-scale policy learning~\cite{wang2025scalingcrl}. Here, SAC-HER serves only as an additional reference for conventional hindsight-based goal-conditioned RL in the same failure-terminated setting. This comparison is not intended to test the CRL-specific missing-survival-mass mechanism identified in Section \ref{sec: gap}. As shown in Figure~\ref{fig:sac_her}, SAC-HER performs substantially worse than Safe-CRL on the four representative failure-prone tasks, particularly on the more challenging locomotion environments.

\section{Supplementary results and ablations}
\label{app:supplementary}

This appendix provides additional quantitative results and analyses that complement the main experiments.

\paragraph{Quantitative results for the main benchmark}

Table~\ref{tab: main-number} reports the complete numerical results corresponding to Figures~\ref{fig: open-figure} and~\ref{fig: main_survival_coverage}.

\begin{table*}[h!]
\centering
\caption{Final performance on the main benchmark, reported as mean $\pm$ standard deviation over five random seeds.}
\label{tab:main-benchmark-results}
\scriptsize
\setlength{\tabcolsep}{3.5pt}
\begin{tabular}{lcccccc}
\toprule
& \multicolumn{2}{c}{Time at goal}
& \multicolumn{2}{c}{Survival time}
& \multicolumn{2}{c}{Goal coverage (\%)} \\
\cmidrule(lr){2-3}
\cmidrule(lr){4-5}
\cmidrule(lr){6-7}
Environment
& Scaling-CRL & Safe-CRL
& Scaling-CRL & Safe-CRL
& Scaling-CRL & Safe-CRL \\
\midrule
Point Goal
& $1.94 \pm 0.11$ & $\mathbf{1.98 \pm 0.09}$
& $555.1 \pm 12.3$ & $\mathbf{631.5 \pm 21.0}$
& $81.6 \pm 3.4$ & $\mathbf{83.0 \pm 1.1}$ \\

Car Goal
& $2.17 \pm 0.04$ & $\mathbf{2.39 \pm 0.06}$
& $464.9 \pm 6.5$ & $\mathbf{550.1 \pm 15.9}$
& $74.2 \pm 1.9$ & $\mathbf{79.7 \pm 0.7}$ \\

Ant Goal
& $276.83 \pm 12.81$ & $\mathbf{684.00 \pm 14.57}$
& $422.1 \pm 16.2$ & $\mathbf{910.8 \pm 8.0}$
& $\mathbf{87.2 \pm 0.9}$ & $81.0 \pm 1.6$ \\

Humanoid Goal
& $22.19 \pm 2.64$ & $\mathbf{266.43 \pm 20.70}$
& $145.7 \pm 12.9$ & $\mathbf{386.3 \pm 19.9}$
& $60.9 \pm 0.7$ & $\mathbf{78.5 \pm 0.9}$ \\

Ant H-Maze
& $390.60 \pm 52.80$ & $\mathbf{412.96 \pm 17.92}$
& $586.6 \pm 66.6$ & $\mathbf{795.6 \pm 19.6}$
& $\mathbf{58.3 \pm 4.6}$ & $54.0 \pm 2.5$ \\

Ant Cross Maze
& $215.55 \pm 72.07$ & $\mathbf{314.46 \pm 27.89}$
& $425.6 \pm 92.7$ & $\mathbf{785.2 \pm 5.6}$
& $\mathbf{56.1 \pm 1.2}$ & $46.0 \pm 3.4$ \\

Ant Big Maze
& $122.12 \pm 36.23$ & $\mathbf{370.51 \pm 14.73}$
& $300.2 \pm 29.0$ & $\mathbf{768.4 \pm 11.3}$
& $\mathbf{62.4 \pm 2.5}$ & $51.8 \pm 1.7$ \\

Humanoid U-Maze
& $0.56 \pm 0.56$ & $\mathbf{200.29 \pm 82.31}$
& $130.4 \pm 42.4$ & $\mathbf{373.9 \pm 102.0}$
& $5.6 \pm 5.6$ & $\mathbf{31.9 \pm 10.3}$ \\

Ant Big Pitfall
& $116.84 \pm 5.23$ & $\mathbf{452.03 \pm 42.61}$
& $294.9 \pm 9.4$ & $\mathbf{821.1 \pm 16.1}$
& $\mathbf{72.9 \pm 5.1}$ & $62.7 \pm 4.6$ \\

Ant Hardest Pitfall
& $108.54 \pm 11.02$ & $\mathbf{348.07 \pm 56.75}$
& $491.0 \pm 6.7$ & $\mathbf{846.9 \pm 15.7}$
& $41.8 \pm 2.4$ & $\mathbf{51.3 \pm 8.8}$ \\

Humanoid Loop Pitfall
& $14.18 \pm 2.43$ & $\mathbf{492.95 \pm 19.91}$
& $120.5 \pm 22.3$ & $\mathbf{637.9 \pm 18.7}$
& $53.7 \pm 19.3$ & $\mathbf{68.0 \pm 1.7}$ \\

Humanoid Big Pitfall
& $8.21 \pm 3.69$ & $\mathbf{474.12 \pm 18.91}$
& $111.8 \pm 23.0$ & $\mathbf{646.5 \pm 17.4}$
& $30.8 \pm 14.1$ & $\mathbf{66.7 \pm 1.9}$ \\
\bottomrule
\label{tab: main-number}
\end{tabular}
\end{table*}

\paragraph{Effect of goal respawning}

\begin{figure}[h!]
\begin{center}
%\framebox[4.0in]{$\;$}
\includegraphics[width=0.53\columnwidth]{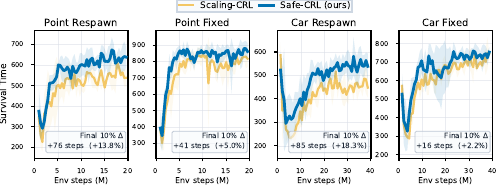}
\end{center}
\caption{Comparison of survival time with respawned and fixed goals in Point Goal and Car Goal. Percentages show the mean advantage of Safe-CRL over Scaling-CRL during the final 10\% of training.}
\label{fig:respawn_ablation}
\end{figure}

Point Goal and Car Goal respawn a new commanded goal after the previous goal is reached, so the policy must continue acting and can still encounter failure afterwards. We compare this default setting with a fixed-goal variant in Figure~\ref{fig:respawn_ablation}. Safe-CRL retains an advantage in both settings, but the improvement in survival time is larger when goals respawn: $+13.8\%$ versus $+5.0\%$ for Point Goal and $+18.3\%$ versus $+2.2\%$ for Car Goal. This result is consistent with the survival-mass interpretation: when control continues after reaching a goal, failure removes additional future occupancy and therefore has a larger effect on subsequent goal-conditioned behaviour.

\paragraph{Z-encoder depth}

We compare 4-layer and 64-layer Z-encoders on four representative tasks while keeping the actor and critic depths fixed. Increasing the Z-encoder depth provides no consistent improvement in either time at goal or survival time, suggesting that a lightweight estimator is sufficient in these experiments.

\begin{figure}[h!]
\begin{center}
\includegraphics[width=0.49\columnwidth]{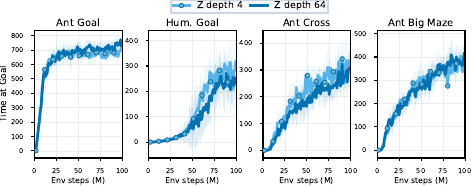}
\includegraphics[width=0.49\columnwidth]{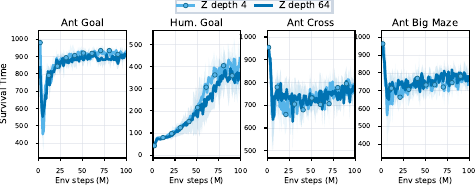}
\end{center}
\caption{Ablation on Z-encoder depth. Actor and critic depths are fixed at 16 layers.}
\label{fig:z-depth-ablation}
\end{figure}

\paragraph{TD-based survival-mass estimation}

Safe-CRL trains the Z-encoder from $Y_t=\mathbf{1}\{T_f>H_t\}$ with $H_t\sim q_\gamma$. We additionally examine whether $Z(s,a)$ can instead be learned through temporal-difference (TD) learning. We replace the Z-encoder objective with a clipped double-Q formulation~\cite{fujimoto2018td3} that estimates a goal-independent survival-mass function.

\begin{figure}[h!]
\begin{center}
%\framebox[4.0in]{$\;$}
\includegraphics[width=0.49\columnwidth]{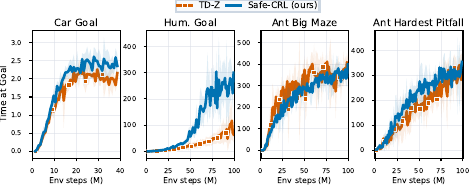}
\includegraphics[width=0.49\columnwidth]{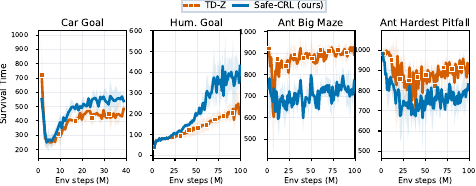}
\end{center}
\caption{Ablations on TD-learned Z. Actor and critic depths are set to 16 layers (8 for Car Goal).}
\label{fig: td ablations}
\end{figure}

As shown in Figure~\ref{fig: td ablations}, TD-based estimation produces mixed results. It substantially degrades goal-reaching and survival performance in Car Goal and Humanoid Goal, while remaining competitive on the selected Ant tasks. We therefore use direct Monte Carlo survival-mass estimation as the default because it is simpler and provides more consistent performance across the evaluated environments.

\paragraph{Computational overhead}

Table~\ref{tab:training-time} reports the wall-clock training time. Humanoid locomotion tasks are tested on modified NVIDIA RTX 4090 (48 GB) and the others are tested on modified NVIDIA RTX 4080 (32 GB). For each environment, Scaling-CRL and Safe-CRL are measured on the same hardware configuration. With 64-layer actor and critic networks, adding the 4-layer Z-encoder increases the average training time from $9.07$ to $9.32$ hours per 100M environment steps, corresponding to an average overhead of only $2.8\%$.

\begin{table}[h!]
\centering
\caption{Average wall-clock training time per 100M environment steps.}
\label{tab:training-time}
\begin{tabular}{lcc}
\toprule
Environment & Scaling-CRL & Safe-CRL (ours) \\
\midrule
Point Goal               &  2.92 h &  3.07 h \\
Car Goal                 &  8.73 h &  8.94 h \\
Ant Goal                 &  6.65 h &  6.71 h \\
Humanoid Goal            & 7.28 h &  7.74 h \\
Ant H-Maze               & 10.20 h & 10.58 h \\
Ant Cross Maze           & 10.48 h & 10.64 h \\
Ant Big Maze             & 10.49 h & 10.70 h \\
Humanoid U-Maze          & 11.09 h & 11.13 h \\
Ant Big Pitfall          &  9.85 h & 10.01 h \\
Ant Hardest Pitfall      & 10.42 h & 10.66 h \\
Humanoid Loop Pitfall    & 10.55 h & 11.01 h \\
Humanoid Big Pitfall     & 10.21 h & 10.70 h \\
\midrule
Mean                     & 9.07 h & \textbf{9.32 h ($\uparrow$ 2.8\%)} \\
\bottomrule
\end{tabular}
\end{table}

\paragraph{Exploration remains a limitation}

\begin{figure}[h!]
\begin{center}
%\framebox[4.0in]{$\;$}
\includegraphics[height=0.15\columnwidth]{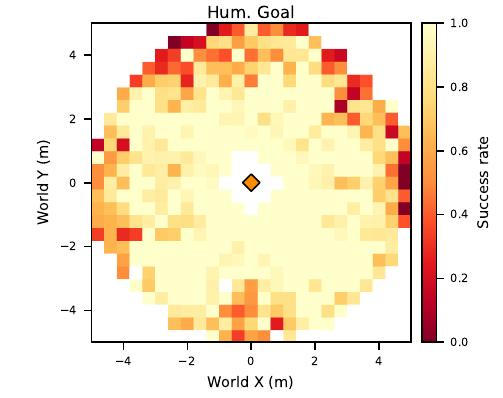}
\includegraphics[height=0.15\columnwidth]{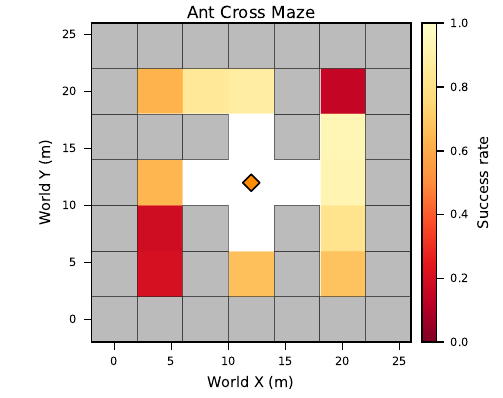}
\includegraphics[height=0.15\columnwidth]{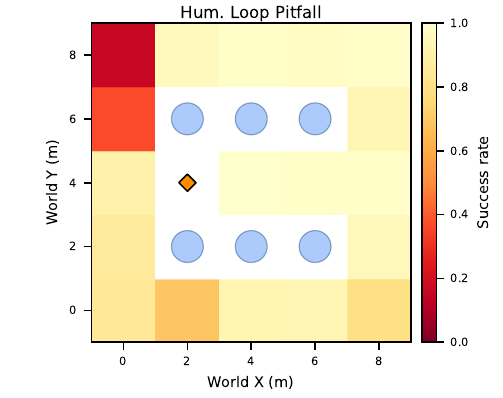}
\includegraphics[height=0.15\columnwidth]{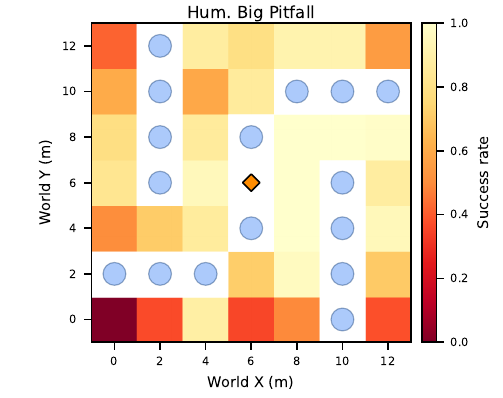}
\includegraphics[height=0.15\columnwidth]{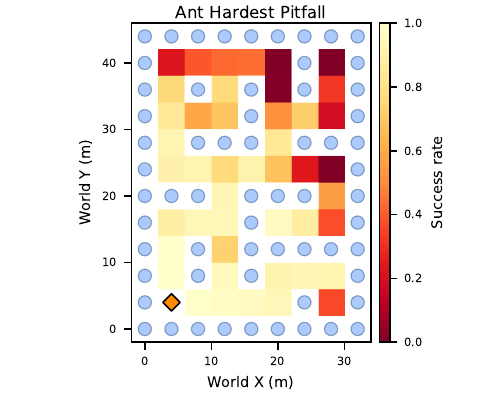}
\end{center}
\caption{Success rate for goals at different places in the playground. Darker colors indicate lower goal-reaching success rates. The robot initially faces the +x axis direction ($\rightarrow$). The diamond marks the initial position of the robot.}
\label{fig: limit-on-explore}
\end{figure}

Safe-CRL corrects the occupancy information used for critic and policy learning, but does not modify the exploration mechanism of CRL. Figure~\ref{fig: limit-on-explore} visualizes goal-reaching success rates across different goal locations. The learned success distributions can remain spatially uneven, including asymmetric performance at geometrically similar goal locations. For example, for Hum. Loop Pitfall, the top-left corner and the bottom-left corner represent the left-back and right-back side of the humanoid robot, respectively. The right-back side (with lighter colors) is better explored and learned than the left-back side. This pattern indicates non-uniform exploration and shows that restoring survival mass does not by itself guarantee uniform goal-space coverage. If the robot's initial position is near a corner, then exploration is restricted. The success rate therefore depends on the distance between the robot and the goal.

\end{document}